\documentclass{article}

\usepackage{bm}

\usepackage[main,preprint]{neurips_2026}
\usepackage{amsmath}
\usepackage{graphicx}
\usepackage{booktabs}
\usepackage{multirow}
\usepackage{caption}

\usepackage[utf8]{inputenc}
\usepackage[T1]{fontenc}
\usepackage{hyperref}
\usepackage{url}
\usepackage{amsfonts}
\usepackage{microtype}
\usepackage{xcolor}

\title{ProtoVis-Nav: Prototypical Visual Imagination for Visual-Spatial Aligned UAV Navigation}

\author{Guangyuan Sui \qquad Yunzhong Hou\thanks{Corresponding author: \texttt{yunzhong.hou@bit.edu.cn}.} \qquad Chi Harold Liu \\
  Beijing Institute of Technology \\
  \texttt{\{3120255778,yunzhong.hou,chiliu\}@bit.edu.cn}
}

\begin{document}

\maketitle

\begin{abstract}
Vision language navigation (VLN) is very challenging for Unmanned Aerial Vehicles (UAVs) since small errors in 3D motion prediction can lead to navigation errors and collisions. Unable to fathom execution outcomes in 3D environments, existing methods often struggle to predict accurate navigation waypoints from language instructions. To solve this, we introduce the Prototypical Visual Imagination framework, shortened as ProtoVis-Nav, which forces the model to imagine an outcome-associated local visual representation for the intended next UAV movement. Specifically, we predict feature prototypes and use their linear combination to estimate the visual feature of the target region. In this manner, the VLN model first imagines the local visual representation of the next intended location, and then uses predicted prototypes as context for generating the final action, promoting alignment between visual context and waypoint outputs. Extensive experiments demonstrate the effectiveness of the ProtoVis-Nav framework and show substantial improvements over existing baselines. Data and code are available at \href{https://github.com/ygs0002/protovis-nav}{link}.
\end{abstract}

\section{Introduction}
Vision-Language Navigation (VLN) is a challenging problem in embodied AI, enabling agents to navigate complex environments while following natural language instructions \cite{gu2022vision,park2023visual}. Although early VLN research predominantly centered on indoor navigation for ground-based robots \cite{krantz2020beyond,ku2020room,zhou2024navgpt,huang2023visual}, a growing body of recent work focuses on Unmanned Aerial Vehicles (UAVs) \cite{cai2025flightgpt,liu2023aerialvln,wu2025aeroduo,zhang2025grounded,xiao2025uav}, 
where navigation involves a more challenging 3D action space and requires precise motion control and reliable 3D trajectory planning \cite{wang2024towards}.
Despite the rapid progress of Vision-Language Models (VLMs) \cite{bai2025qwen3vl,wang2025internvl3,hong2025glm,team2026kimi}, applying them to UAV navigation remains difficult as they struggle to navigate 3D environments accurately and safely \cite{wang2025uavflow,xiao2025uav}.

A major reason for these navigation errors is the inability to associate motion predictions with their potential outcomes \cite{nie2025wmnav}.
In fact, most methods directly predict continuous waypoints from visual observations and language instructions \cite{arafat2023vision}.
Since their training objective penalizes only the waypoint error, the model can hardly anticipate the execution outcomes of its own actions \cite{Koh_2021_ICCV}, increasing the risk of generating trajectories that are potentially unsafe in the 3D environment.

To address this problem, an emerging trend in visual language navigation is to equip the agent with future-aware visual predictive capabilities, often framed as navigation world models \cite{bar2025navigation}. 
Specifically, they usually predict the future observation via a specific video generation model \cite{Koh_2021_ICCV, bar2025navigation} or as outputs of the same VLN model \cite{yao2025navmorph, huang2026unemo, Wang2026monodream}. UAV navigation requires accurate fine-grained motion control grounded in local visual evidence \cite{ding2026history}. In fact, \textit{precise} flight actions are often determined by small and localized visual regions (\textit{e.g.}, a narrow gap to fly through). This raises a key question: rather than generating the global future scene, can we enforce future-aware grounding by strictly focusing on these localized visual regions most critical to the upcoming movement?

In this work, we propose the Prototypical Visual Imagination framework for UAV navigation, dubbed as ProtoVis-Nav. The method is motivated by a simple idea: rather than modeling an entire future scene or a long-horizon latent rollout, the agent should focus on the exact regions most relevant to its precise movement by imagining an outcome-associated local visual representation.
To this end, we project the next intended UAV location onto the current image plane, crop the surrounding local region, and use its visual feature as the supervision target. Compared with using the entire future frame or the future-frame center, this projected region provides a supervision target that more directly grounds 3D motion prediction in the current image space, while avoiding the intractable imagination of future observations.
Plus, we introduce a prototype-based feature prediction method. Instead of using multi-layer perceptrons to directly estimate the target feature, this approach predicts a set of Prototypical Visual Imagination (PVI) tokens, which are then combined through a softmax-normalized mixing matrix, so that their linear combination matches the target-region feature. The predicted PVI tokens are less susceptible to overfitting and are then used as context for waypoint prediction, encouraging tighter alignment between visual context and continuous action outputs.

We experimentally verify the effectiveness of the proposed method on OpenUAV \cite{wang2024towards} and HUGE-Bench \cite{guo2026hugebench}. Results show that ProtoVis-Nav substantially improves navigation performance over strong baselines and leads to more reliable continuous UAV navigation.

\section{Related Works}

\textbf{Vision language navigation for UAVs.}
Vision-Language Navigation (VLN) has traditionally focused on guiding ground agents through structured, near-planar indoor environments \cite{krantz2020beyond,ku2020room,zhou2024navgpt,huang2023visual,zhao2022target,10495141,zhang-etal-2025-mapnav}. Recently, the field has naturally extended to Unmanned Aerial Vehicles (UAVs). UAV-based VLN requires agents to ground language instructions into complex, continuous flight dynamics, demanding a stronger understanding of 3D spatial reasoning. Several realistic and continuous UAV VLN benchmarks have recently emerged \cite{wu2025aeroduo,xiao2025uav,wang2024towards,wang2025uavflow}, providing an essential foundation for open-world UAV VLN research. Building upon them, TravelUAV \cite{wang2024towards} directly regresses the next waypoint based on current observations and UAV kinematics. NavFoM \cite{zhang2025embodied} further explores a cross-embodiment and cross-task navigation foundation model trained on large-scale navigation data, showing strong generalization across different navigation tasks and robotic platforms, including UAV navigation. To better process historical context, LongFly \cite{jiang2025longfly} introduces a history-aware spatiotemporal modeling strategy that transforms fragmented and redundant historical data into structured, compact, and expressive representations. To improve policy learning, OpenVLN \cite{lin2025openvln} formulates a reinforcement learning framework to optimize VLMs for UAV navigation tasks. It efficiently fine-tunes the VLM using rule-based policies under limited training data, and incorporates a long-horizon trajectory planner that generates precise UAV actions via value-based rewards.

Despite these advancements, most continuous UAV VLN methods still mainly rely on current observations, historical context, or geometric priors for waypoint prediction.
Since their training objective focuses on waypoint error, the model receives limited guidance for identifying visual cues associated with the next intended location of the UAV, increasing the risk of generating trajectories that are potentially unsafe in the 3D environment.

\textbf{Navigation world models.}
An emerging trend in visual-language navigation is to equip agents with future-aware visual predictive capabilities, often framed as navigation world models. For example, Pathdreamer studies visual world modeling for indoor navigation, generating future 360$^\circ$ RGB, depth, and semantic observations along hypothetical trajectories with a two-stage structure-to-image generator \cite{Koh_2021_ICCV}. Dreamwalker further adapts this idea to VLN-CE by constructing an explicit environment graph and using a scene synthesizer to imagine future panoramic observations, which are then evaluated through mental planning with Monte Carlo Tree Search \cite{Wang_2023_ICCV}. HNR-VLN predicts multi-level CLIP-based semantic features for candidate future viewpoints via a hierarchical neural radiance representation, enabling more efficient path evaluation \cite{Wang_2024_CVPR}. NWM scales this paradigm to video generation, where a diffusion transformer simulates future visual trajectories conditioned on actions and ranks or plans routes according to goal similarity \cite{bar2025navigation}. 

More recent methods emphasize compact and adaptive latent prediction: UNeMo predicts the next visual state from visual features, language instructions, and candidate actions, feeding the predicted state back to refine navigation decisions \cite{huang2026unemo}. MonoDream uses latent panoramic dreaming to predict current and future panoramic RGB-D features from a forward-facing RGB view, thereby compensating for limited field of view without explicit panoramic sensors \cite{Wang2026monodream}. DriveVLA-W0 uses future-image prediction to supervise driving representations \cite{li2025drivevlaw0}, while ANWM predicts action-conditioned future observations for UAV trajectory evaluation \cite{zhang2025anwm}.

While effective for future-aware visual world models, they focus on generating global scene representations, such as full-frame videos or panoramic latents. However, UAVs require accurate fine-grained motion control, grounded in local visual evidence. Predicting the background and surroundings can provide general orientation, but \textit{precise} flight actions are often determined by small and localized visual regions (\textit{e.g.}, a narrow gap to fly through).
To bridge this gap, our proposed ProtoVis-Nav framework departs from the global imagination paradigm. Instead of predicting complete future frames or global features, we explicitly project the intended future 3D location onto the current 2D image plane and predict prototypical visual tokens strictly for that local region. This localized, prototype-based imagination encourages the UAV to ground its motion decisions in the most critical visual context for more accurate and reliable continuous aerial navigation in 3D space.

Regarding local visual supervision, our method differs from ReconVLA~\cite{song2025reconvla}, which reconstructs gaze regions around the objects to be manipulated in the current image with a diffusion model to guide visual attention. ProtoVis-Nav constructs its local target by projecting the next intended 3D UAV location onto the current 2D image plane, linking local visual supervision to the intended execution outcome of the next action.

\vskip -4pt
\section{ProtoVis-Nav}

\subsection{Overall Pipeline}
\begin{figure}[t]
    \centering
    \includegraphics[width=\linewidth]{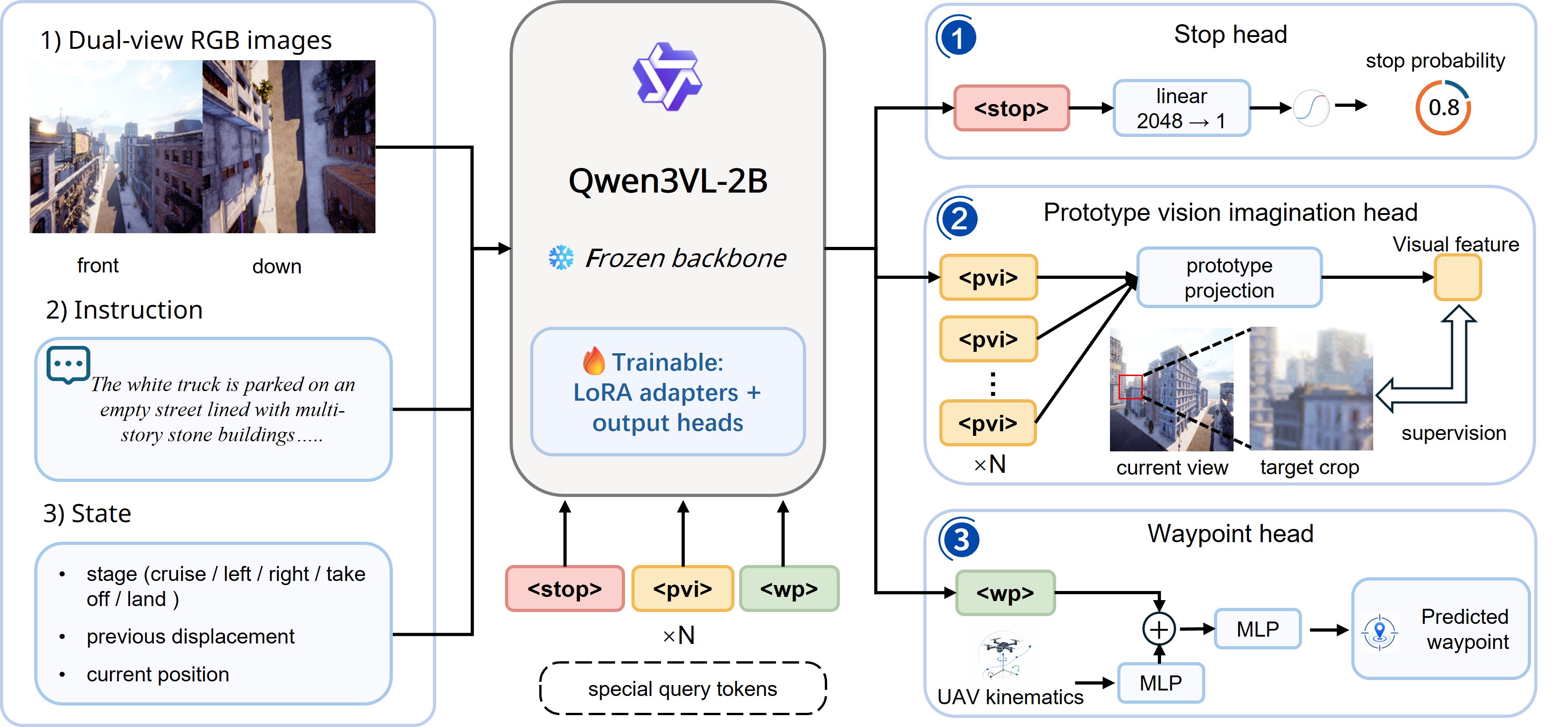}
    \caption{Overall pipeline of ProtoVis-Nav.}
    \label{fig:pipeline}
    \vskip -4pt
\end{figure}

As shown in Fig.~\ref{fig:pipeline}, ProtoVis-Nav uses Qwen3VL-2B-Instruct \cite{bai2025qwen3vl} as the VLM backbone. The input consists of dual-view RGB observations (front-view and down-view images), the language instruction, and the current UAV state, including the previous displacement and current position. The Qwen3VL backbone is kept frozen, while the LoRA adapters inserted into the backbone and the newly added output heads are trained.

We adopt three types of special query tokens to the model input: \texttt{<stop>}, \texttt{<pvi>}, and \texttt{<wp>}. The \texttt{<stop>} token predicts whether the UAV should terminate the current navigation episode. The \texttt{<pvi>} tokens are short for Prototypical Visual Imagination (PVI), which produce visual prototypes that capture the local visual features around the next intended UAV location. These PVI tokens are then used as context for waypoint prediction. The hidden state of the \texttt{<wp>} token is used by the waypoint head to predict a waypoint for short-horizon trajectory generation.

Specifically, the hidden state of \texttt{<stop>} is passed through a linear layer and a sigmoid activation to produce a stop probability $p_{\mathrm{stop}}$. When $p_{\mathrm{stop}}$ exceeds a predefined threshold $\tau_{\mathrm{stop}}$, the model issues a stop signal. Details of stop-head supervision are provided in Appendix~\ref{secsec:stop head}. 

For visual imagination, the $K$ \texttt{<pvi>} tokens predict the feature prototypes, which are combined through a softmax-normalized linear weighting to produce an outcome-associated local visual representation of the target local region. This encourages the model to anticipate what the relevant visual context should look like before predicting the next action. See Section~\ref{secsec:prototype head}.

Finally, the \texttt{<wp>} token considers the \texttt{<pvi>} tokens as context within the VLM, and the waypoint head combines its hidden state with UAV kinematic information to predict the next waypoint.
Details of the waypoint head are in the Appendix~\ref{secsec:waypoint head}. 

\begin{figure}[t]
    \centering
    \vskip -2mm
    \includegraphics[width=\linewidth]{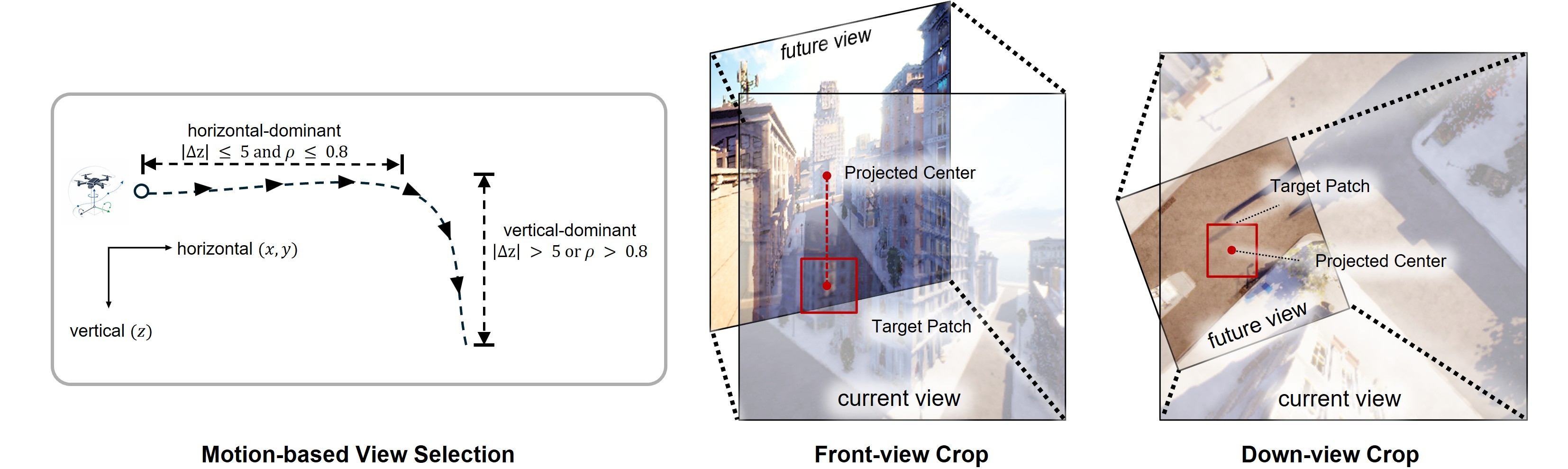}
    \vskip -1mm
    \caption{Illustration of target-region visual feature selection during training. The left panel shows our motion-dominance analysis for selecting the front or down view. Given the selected view, the next intended UAV location is projected onto the current image plane, and a local target patch is cropped around the projected center to supervise prototypical visual imagination.}
    \label{fig:target_selection}
    \vskip -3mm
\end{figure}

\subsection{Visual Imagination Target}

During training, for the current frame $t$ and the future frame $t+1$, we obtain their UAV positions $\bm{p}_t, \bm{p}_{t+1}\in\mathbb{R}^3$ and orientations $R_t, R_{t+1}\in\mathbb{R}^{3\times3}$ from the recorded expert trajectory.
We also record the frontal and downward duo-views as $\bm{x}_{t}^\text{front} \in \mathbb{R}^{3\times 256\times256}$ and $\bm{x}_{t}^\text{down} \in \mathbb{R}^{3\times 256\times256}$, both of which are of a $256\times 256$ spatial resolution.

For visual imagination supervision, the target-region is a local visual patch in the current observation $\bm{x}_{t}\in\left\{\bm{x}_{t}^\text{front},\bm{x}_{t}^\text{down}\right\}$, which provides an outcome-associated local visual representation of the next intended UAV location.
The entire future frame mostly describes background regions that are not directly related to the precise 3D motions, and predicting the future frame would introduce an intractable prediction.
Thus, we do not use the future observation $\bm{x}_{t+1}\in\left\{\bm{x}_{t+1}^\text{front},\bm{x}_{t+1}^\text{down}\right\}$, either as an entire frame or using its center, for the imagination target.
Instead, we only focus on a small patch from the current observation that corresponds to the future UAV location to visually ground the 3D waypoint.
In this design, the VLM essentially only needs to point out the future location from the current observation, offering a robust supervision target.

To locate this region from the current observation, we use the next drone location $\bm{p}_{t+1}$ and project this point onto the current camera image plane,
\begin{equation}
    \bm{p}_\text{proj}
    =
    \bm{R}_{t}^{\top}
    (\bm{p}_{t+1} - \bm{p}_{t}),
\end{equation}
where $\bm{p}_\text{proj}$ denotes the projected location in the current UAV reference frame. 
We then crop a $32\times 32$ patch around the projected coordinate,
\begin{equation}
\label{eq:patch}
\bm{\Tilde{x}}=\texttt{patch}\left(\bm{x}_t, \bm{p}_\text{proj}\right) ,
\end{equation}
where  $\bm{\Tilde{x}} \in \mathbb{R}^{3\times 32\times 32}$ denotes the $32\times 32$ image patch from $\bm{x}_t$. 
The visual feature of this local patch is then used as the supervision target for prototypical visual imagination. 

Before we can crop out the target region $\bm{\Tilde{x}}$, one remaining issue is to determine which one of the duo-view observations to use. While both front and down views are fed into the VLM (see Fig.~\ref{fig:pipeline}), only one of them best reflects the intended future location. We use a simple intuition to select the view: front view is more suitable for forward motion, while down view is more informative for take-off, landing, or other vertical-dominant movements. 
Specifically, we consider two cues. 

\textit{First}, we compute the vertical-to-horizontal movement ratio,
\begin{equation}
    \rho = \frac{\|\bm{p}_{\mathrm{proj}}^{z}\|_2}{\| \bm{p}_{\mathrm{proj}}^{xy} \|_2},
\end{equation}
where $\bm{p}_{\mathrm{proj}}^{xy} \in \mathbb{R}^2$ and $\bm{p}_{\mathrm{proj}}^{z} \in \mathbb{R}$ denote the horizontal and vertical displacements.

\textit{Second}, we check whether the UAV is clearly ascending or descending in a short temporal window around the current step. Specifically, we compare the altitude before the current step with the altitude a few steps ahead, 
\begin{equation}
    \Delta z = z_{t+2} - z_{t-1}.
\end{equation}
A positive $\Delta z$ indicates ascending motion, while a negative value indicates descending motion. We use its absolute value to cover both cases. 

\textit{Overall}, the motion is regarded as vertical-dominant if the short-window altitude change satisfies $|\Delta z|>5$ or the body-frame vertical-to-horizontal ratio satisfies $\rho>0.8$; otherwise, it is regarded as horizontal-dominant. 
For vertical-dominant motion, the target patch is extracted from the down view, while for horizontal-dominant motion, it is extracted from the front view:
\begin{equation}
    \bm{x}_t=
    \begin{cases}
    \bm{x}_t^\text{down}, & \text{if } 
    |\Delta z| > 5
    \ \text{or}\
    \rho > 0.8
    , \\
    \bm{x}_t^\text{front}, & \text{otherwise}.
    \end{cases}
\end{equation}
where $\bm{x}_t$ determines the duo-view selection in Eq.~\ref{eq:patch}.  Fig.~\ref{fig:target_selection} illustrates the motion-based view selection and the target-region construction. 
The left panel shows how horizontal-dominant and vertical-dominant motions determine which view is used.
After selecting the view, the next intended UAV location is projected onto the current image plane, and a local target patch is cropped around the projected center as the visual imagination target, as shown in the right panel.

\subsection{Prototypical Visual Imagination Head}
\label{secsec:prototype head}

\begin{figure}[t]
    \centering
    \includegraphics[width=\linewidth]{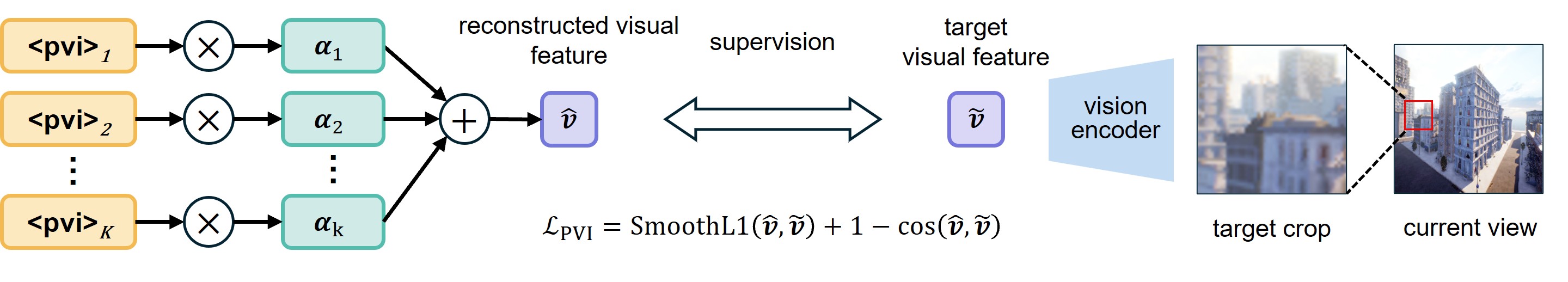}
    \caption{Illustration of the prototypical visual imagination head with $K$ \texttt{<pvi>} tokens and one output visual token. The hidden states of multiple \texttt{<pvi>} tokens are combined by softmax-normalized weights into a predicted target-region visual feature, which is supervised by the target crop feature extracted from the visual encoder.}
    \label{fig:pvi_head}
\end{figure}

As shown in Fig.~\ref{fig:pvi_head}, the Prototypical Visual Imagination Head is a lightweight module that treats the \texttt{<pvi>} token outputs as visual feature prototypes \cite{snell2017prototypical}. This module learns a linear combination of these prototypes to match the target-region visual feature. This design uses the \texttt{<pvi>} token outputs as a set of bases for encoding the target-region visual cues, which works better at providing context for the \texttt{<wp>} prediction. We verify this empirically and show that our prototype design outperforms the MLP alternative, see Section~\ref{sec:ablation}. 

We consider $K=16$  \texttt{<pvi>} tokens $\bm{h}_k,\ k\in\left\{1,\ldots,K\right\}$. Using a learned softmax mixing matrix,
\begin{equation}
    \bm{A}=\{a_{n,k}\}, \quad \alpha_{n,k}
    =
    \frac{\exp(a_{n,k})}
    {\sum_{j=1}^{K}\exp(a_{n,j})},
\end{equation}
where  $N$ is the number of target visual tokens for describing $\bm{\Tilde{x}}$ in Eq.~\ref{eq:patch}, $\alpha_{n,k}$ denotes the mixing weight from the $k$-th PVI token to the $n$-th visual token, $\bm{A}\in\mathbb{R}^{N\times K}$ is the overall mixing logits. The predicted visual tokens are obtained by
\begin{equation}
    \bm{\hat{v}}_n
    =
    \sum_{k=1}^{K}
    \alpha_{n,k}\bm{h}_k,
    \quad n=\left\{1,\ldots,N\right\}.
\end{equation}
For the $32\times32$ target crop, the Qwen3-VL visual encoder produces $N=1$, so the head represents the target-region visual feature as a soft combination of 16 PVI tokens.

The supervision target is the feature extracted by the Qwen3VL visual encoder,
\begin{equation}
    \left\{\bm{\Tilde{v}}_n\right\}
    =
    \texttt{encoder}
    (\bm{\Tilde{x}}), 
    \quad
    n=\left\{1,\ldots,N\right\}.
\end{equation} 
The predicted features $\left\{\bm{\hat{v}}\right\}$ are supervised to match the target features $\left\{\bm{\Tilde{v}}\right\}$ using this PVI loss, 
\begin{equation}
    \mathcal{L}_\text{PVI}
    =
    \frac{1}{N}
    \sum_{n=1}^{N}
    \left(
    \mathrm{SmoothL1}(\bm{\hat{v}}_n,\bm{\Tilde{v}}_n)
    +
    1 -
    \cos(\bm{\hat{v}}_n,\bm{\Tilde{v}}_n)
    \right).
\end{equation}
Through this supervision, the \texttt{<pvi>} tokens are encouraged to encode the visual feature of the local region around the next intended UAV location, and their hidden states provide context for the waypoint token within the VLM.

\begingroup
\setlength{\floatsep}{12pt plus 2pt minus 2pt}
\setlength{\textfloatsep}{16pt plus 2pt minus 2pt}
\section{Experiments}

\subsection{Experimental Setup}
\label{sec:exp_setup}

\textbf{Benchmarks and platform.}
We conduct experiments on OpenUAV and the obstacle-aware region traversal tasks in HUGE-Bench. OpenUAV is built on the TRAVEL platform \cite{wang2024towards}. It provides diverse outdoor environments, realistic flight dynamics, and closed-loop simulation support for continuous UAV navigation. The platform supports physics-based UAV control with 6-DoF pose representation and configurable sensors such as RGB/depth cameras. The OpenUAV benchmark contains human-operated UAV trajectories paired with multi-view RGB-D observations, language instructions, UAV states, and waypoint annotations. OpenUAV evaluation uses the L1 assistant setting. HUGE-Bench combines 3D Gaussian Splatting with mesh geometry to support photorealistic rendering and collision-aware evaluation of UAV tasks~\cite{guo2026hugebench}.

\textbf{Data split.}
We follow the official data split of OpenUAV. The training set contains 9,152 trajectories. For evaluation, we report results on the Test Seen and Test Unseen splits. 
The Test Seen split contains 1,410 trajectories from objects and scenes seen in the training set. The Test Unseen split contains 1,587 trajectories from maps or objects that do not appear in the training set. Following the official OpenUAV setting, test episodes are further grouped into Full, Easy, and Hard subsets according to trajectory length. Specifically, trajectories shorter than 250 meters are labeled as Easy, while trajectories at or above 250 meters are labeled as Hard. Unless otherwise specified, we report results on the Full subset for variant studies and report Full, Easy, and Hard results for state-of-the-art comparison.
For HUGE-Bench, we use the obstacle-aware region traversal task subset \cite{guo2026hugebench}, which requires traversing a designated area while avoiding obstacles.
We train and evaluate models separately within each benchmark.

\textbf{Evaluation metrics.}
We evaluate OpenUAV navigation performance using four metrics: Navigation Error (NE), Success Rate (SR), Oracle Success Rate (OSR), and Success weighted by Path Length (SPL) \cite{wang2024towards}. 
NE measures the distance between the UAV's final position and the target position, and is reported in meters; a lower NE indicates that the UAV stops closer to the target. 
SR measures the percentage of episodes in which the UAV successfully stops within a predefined distance threshold from the target, and is reported as a percentage. 
Following the official OpenUAV setting, an episode is regarded as successful if the UAV stops within 20 meters of the target. 
OSR measures whether the UAV reaches the target region at any time during navigation, regardless of whether it finally stops there, and is reported as a percentage. 
SPL jointly considers navigation success and path efficiency by weighting successful episodes according to the ratio between the shortest and the executed path length, and is also reported as a percentage. 
For NE, lower is better; for SR, OSR, and SPL, higher is better.
For HUGE-Bench, higher normalized Dynamic Time Warping (nDTW) indicates closer alignment with the expert trajectory. Higher Collision-aware Success weighted by Path Length (CSPL) scores and a lower Collision Rate (CR) indicate better performance.

\textbf{Baselines.}
We compare ProtoVis-Nav with representative UAV VLN baselines on OpenUAV, including Random Action, CMA \cite{anderson2018vision}, TravelUAV \cite{wang2024towards}, NavFoM \cite{zhang2025embodied}, and LongFly \cite{jiang2025longfly}. Random Action serves as a simple lower-bound policy. CMA is a classical cross-modal attention baseline. TravelUAV is the original VLM-based method on the OpenUAV benchmark. NavFoM is a navigation foundation model evaluated on OpenUAV, and LongFly is a recent spatiotemporal context modeling method for long-horizon UAV VLN. We also report Base (Ours), which uses the same VLM backbone, LoRA setting, stop head, and waypoint head as ProtoVis-Nav, but removes the \texttt{<pvi>} tokens and the PVI Head. On HUGE-Bench, we compare ProtoVis-Nav with Base.

\textbf{Implementation details.}
ProtoVis-Nav adopts Qwen3VL-2B-Instruct \cite{bai2025qwen3vl} as the VLM backbone. During training, we freeze the backbone parameters and train only the LoRA adapters and the newly added output heads. ProtoVis-Nav takes $256\times256$ front-view and down-view RGB images, the language instruction, and UAV state information as input on OpenUAV. HUGE-Bench uses front-view RGB images. The UAV pose and waypoint annotations are used during training to construct the projected target-region supervision for Prototypical Visual Imagination. By default, we use $K=16$ \texttt{<pvi>} tokens and a $32\times32$ target crop. The Prototypical Visual Imagination Head uses a softmax-normalized mixing matrix to combine PVI token hidden states into a target-region visual representation for feature supervision. We train ProtoVis-Nav using AdamW with a learning rate of $5\times10^{-5}$ and a global batch size of 64. During evaluation, the maximum number of waypoint steps is set to 200. Training is conducted on 4 NVIDIA RTX 5090 GPUs, and inference is performed on a single NVIDIA RTX A6000 GPU.

\subsection{Comparison with State-of-the-Art Methods}
\label{sec:sota_comparison}

We compare ProtoVis-Nav with representative UAV VLN methods on the OpenUAV benchmark and the obstacle-aware region traversal tasks of HUGE-Bench. 
As shown in Table~\ref{tab:sota_all}, ProtoVis-Nav achieves competitive performance across both seen and unseen evaluation settings.

\begin{table}[t]
\centering
\caption{Comparison with state-of-the-art methods on the OpenUAV benchmark.}
\label{tab:sota_all}

\small
\setlength{\tabcolsep}{3pt}

\textbf{Test Seen}

\resizebox{\linewidth}{!}{
\begin{tabular}{lcccccccccccc}
\toprule
\multirow{2}{*}{Method} 
& \multicolumn{4}{c}{Full} 
& \multicolumn{4}{c}{Easy} 
& \multicolumn{4}{c}{Hard} \\
\cmidrule(lr){2-5} \cmidrule(lr){6-9} \cmidrule(lr){10-13}
& NE $\downarrow$ & SR $\uparrow$ & OSR $\uparrow$ & SPL $\uparrow$
& NE $\downarrow$ & SR $\uparrow$ & OSR $\uparrow$ & SPL $\uparrow$
& NE $\downarrow$ & SR $\uparrow$ & OSR $\uparrow$ & SPL $\uparrow$ \\
\midrule
Random Action & 222.20 & 0.14 & 0.21 & 0.07 & 142.07 & 0.26 & 0.39 & 0.13 & 320.12 & 0.00 & 0.00 & 0.00 \\
CMA           & 135.73 & 8.37 & 18.72 & 7.90 & 84.89 & 11.48 & 24.52 & 10.68 & 197.77 & 4.57 & 11.65 & 4.51 \\
TravelUAV     & 106.28 & 16.10 & 44.26 & 14.30 & 68.78 & 18.84 & 47.61 & 16.39 & 152.04 & 12.76 & 40.16 & 11.76 \\
NavFoM        & 93.05 & 29.17 & 49.24 & 25.03 & 58.98 & 32.91 & 53.16 & 27.87 & 143.83 & 23.58 & 43.40 & 20.80 \\
LongFly       & 60.02 & 36.39 & 65.87 & 31.07 & 38.10 & 38.52 & \textbf{71.90} & 31.24 & 85.20 & 33.94 & 58.94 & 30.88 \\
\midrule
Base   & 55.67 & 46.47 & 62.69 & 41.14 & 42.46 & 49.04 & 62.71 & 41.94 & \textbf{71.96} & 43.31 & 62.68 & 40.14 \\
ProtoVis-Nav & \textbf{51.77} & \textbf{50.78} & \textbf{66.50} & \textbf{44.68} & \textbf{35.13} & \textbf{54.92} & 68.84 & \textbf{46.68} & 72.28 & \textbf{45.67} & \textbf{63.62} & \textbf{42.21} \\
\bottomrule
\end{tabular}
}

\vspace{2mm}

\textbf{Test Unseen}

\resizebox{\linewidth}{!}{
\begin{tabular}{lcccccccccccc}
\toprule
\multirow{2}{*}{Method} 
& \multicolumn{4}{c}{Full} 
& \multicolumn{4}{c}{Easy} 
& \multicolumn{4}{c}{Hard} \\
\cmidrule(lr){2-5} \cmidrule(lr){6-9} \cmidrule(lr){10-13}
& NE $\downarrow$ & SR $\uparrow$ & OSR $\uparrow$ & SPL $\uparrow$
& NE $\downarrow$ & SR $\uparrow$ & OSR $\uparrow$ & SPL $\uparrow$
& NE $\downarrow$ & SR $\uparrow$ & OSR $\uparrow$ & SPL $\uparrow$ \\
\midrule
Random Action & 225.64 & 0.06 & 0.06 & 0.06 & 164.66 & 0.19 & 0.19 & 0.19 & 280.58 & 0.00 & 0.00 & 0.00 \\
CMA           & 147.27 & 4.98 & 12.41 & 4.74 & 102.54 & 8.03 & 17.52 & 7.52 & 191.30 & 2.76 & 7.53 & 2.71 \\
TravelUAV     & 130.60 & 11.41 & 31.13 & 10.45 & 96.27 & 12.47 & 33.31 & 11.29 & 167.49 & 10.62 & 28.91 & 9.80 \\
NavFoM        & 118.34 & 15.63 & 30.46 & 14.21 & 89.77 & 16.98 & 32.22 & 15.35 & 155.69 & 14.35 & 27.79 & 13.16 \\
LongFly       & 91.84 & 24.19 & \textbf{43.86} & 20.84 & \textbf{69.16} & 22.89 & \textbf{43.24} & 18.66 & 112.02 & 25.36 & 44.41 & 22.76 \\
\midrule
Base   & 98.04 & 27.91 & 40.71 & 25.30 & 78.25 & 27.92 & 37.01 & 24.63 & 116.70 & 27.91 & 44.19 & 25.92 \\
ProtoVis-Nav & \textbf{90.85} & \textbf{30.12} & 41.52 & \textbf{27.53} & 75.43 & \textbf{29.22} & 38.44 & \textbf{26.06} & \textbf{105.39} & \textbf{30.97} & \textbf{44.43} & \textbf{28.92} \\
\bottomrule
\end{tabular}
}
\end{table}

\textbf{Results on OpenUAV Test Seen.}
\label{sec:test_seen}
On the Test Seen split, ProtoVis-Nav obtains strong results across the Full, Easy, and Hard subsets. 
For the Full subset, ProtoVis-Nav achieves an NE of $51.77$m, an SR of $50.78\%$, an OSR of $66.50\%$, and an SPL of $44.68\%$, outperforming all compared methods on the four metrics. 
These results support our motivation that predicting a localized target-region visual feature can provide useful future-aware grounding for continuous waypoint prediction.

On the Easy subset, ProtoVis-Nav achieves an NE of $35.13$m, an SR of $54.92\%$, an OSR of $68.84\%$, and an SPL of $46.68\%$. 
It obtains the best NE, SR, and SPL, while LongFly achieves the best OSR. 
On the Hard subset, ProtoVis-Nav obtains an NE of $72.28$m, an SR of $45.67\%$, an OSR of $63.62\%$, and an SPL of $42.21\%$. 
It achieves the best SR, OSR, and SPL, while Base (Ours) has a slightly lower NE. 
This shows that ProtoVis-Nav improves success-oriented metrics on longer trajectories, although it does not uniformly improve final-position error in the Hard setting.

Compared with Base (Ours), ProtoVis-Nav brings   consistent gains on the Full subset, reducing NE from $55.67$m to $51.77$m and improving SR from $46.47\%$ to $50.78\%$. 
It also improves OSR from $62.69\%$ to $66.50\%$ and SPL from $41.14\%$ to $44.68\%$. 
These results show that prototypical visual imagination offers a useful signal beyond direct waypoint supervision.

\textbf{Results on OpenUAV Test Unseen.}
\label{sec:test_unseen}
On the Test Unseen split, ProtoVis-Nav achieves competitive generalization performance. 
For the Full subset, ProtoVis-Nav obtains an NE of $90.85$m, an SR of $30.12\%$, an OSR of $41.52\%$, and an SPL of $27.53\%$. 
It achieves the best NE, SR, and SPL among the compared methods, while LongFly obtains a slightly higher OSR. 
The drop from Test Seen to Test Unseen indicates that navigation in unseen maps or with unseen objects remains challenging.

On the Easy subset, ProtoVis-Nav achieves an NE of $75.43$m, an SR of $29.22\%$, an OSR of $38.44\%$, and an SPL of $26.06\%$, obtaining the best SR and SPL. 
On the Hard subset, ProtoVis-Nav achieves an NE of $105.39$m, an SR of $30.97\%$, an OSR of $44.43\%$, and an SPL of $28.92\%$, obtaining the best results on all four metrics. 

Compared with Base (Ours), ProtoVis-Nav improves the Full-set SR from $27.91\%$ to $30.12\%$ and reduces NE from $98.04$m to $90.85$m. 
It also improves OSR from $40.71\%$ to $41.52\%$ and SPL from $25.30\%$ to $27.53\%$. 
These results suggest that PVI supervision remains useful for unseen environments, while generalization in continuous UAV VLN is still difficult.

\textbf{Results on HUGE-Bench Obstacle-Aware Region Traversal Tasks.}
Evaluation results on these tasks vary substantially across runs, so Table~\ref{tab:huge_bench} reports the mean over five independent training-and-evaluation runs. Compared with Base, ProtoVis-Nav increases nDTW from $0.1599$ to $0.1762$, SR from $36.39\%$ to $47.47\%$, and CSPL from $0.3179$ to $0.4161$, and reduces CR from $54.46\%$ to $39.76\%$. These gains indicate that PVI supports goal reaching and collision-aware trajectory execution, as traversal requires waypoint choices that account for nearby obstacles.

\begin{table}[!tb]
\centering
\begin{minipage}[t]{0.43\linewidth}
\vskip 0pt
\centering
\captionsetup{type=table,skip=0pt}
\caption{Comparison on HUGE-Bench obstacle-aware region traversal split.}
\label{tab:huge_bench}
\vskip 4pt
\small
\setlength{\tabcolsep}{2pt}
\resizebox{\linewidth}{!}{
\begin{tabular}{ccccc}
\toprule
Model & nDTW $\uparrow$ & SR (\%) $\uparrow$ & CSPL $\uparrow$ & CR (\%) $\downarrow$ \\
\midrule
Base & 0.1599 & 36.39 & 0.3179 & 54.46 \\
ProtoVis-Nav & \textbf{0.1762} & \textbf{47.47} & \textbf{0.4161} & \textbf{39.76} \\
\midrule
$p$-value & 0.3242 & \textbf{0.0458} & \textbf{0.0309} & \textbf{0.0269} \\
\bottomrule
\end{tabular}
}
\end{minipage}%
\hfill
\begin{minipage}[t]{0.55\linewidth}
\vskip 0pt
\centering
\captionsetup{type=table,skip=0pt}
\caption{Comparison of safety metrics for Base and ProtoVis-Nav on OpenUAV.}
\label{tab:openuav_safety}
\vskip 4pt
\small
\setlength{\tabcolsep}{2pt}
\resizebox{\linewidth}{!}{
\begin{tabular}{cccccc}
\toprule
\multirow{2}{*}{Model}
& \multirow{2}{*}{CR (\%) $\downarrow$}
& \multicolumn{3}{c}{Near-collision rate (\%) $\downarrow$}
& \multirow{2}{*}{Clearance (m) $\uparrow$} \\
\cmidrule(lr){3-5}
& & $<0.5$\,m & $<1.0$\,m & $<1.5$\,m & \\
\midrule
Base & 36.44 & 0.3804 & 0.5912 & 0.7938 & 16.7205 \\
ProtoVis-Nav & \textbf{34.41} & \textbf{0.3746} & \textbf{0.5630} & \textbf{0.7555} & \textbf{16.8192} \\
\bottomrule
\end{tabular}
}
\end{minipage}%
\end{table}

\subsection{Safety Analysis on OpenUAV}
\label{sec:safety_analysis}

On OpenUAV, CR measures the percentage of episodes with a collision. Near-collision rate measures the percentage of frames whose obstacle clearance falls below a given threshold, and mean obstacle clearance measures the average distance to the nearest obstacle. We use clearance thresholds of $0.5$m, $1.0$m, and $1.5$m to assess obstacle proximity at different distances.

As shown in Table~\ref{tab:openuav_safety}, ProtoVis-Nav reduces CR from $36.44\%$ to $34.41\%$ and increases mean obstacle clearance from $16.7205$m to $16.8192$m. At these thresholds, near-collision rates decrease from $0.3804\%$, $0.5912\%$, and $0.7938\%$ for Base to $0.3746\%$, $0.5630\%$, and $0.7555\%$ for ProtoVis-Nav, respectively. These reductions indicate less frequent exposure to nearby obstacles during flight.

\begin{figure}[!htbp]
\centering

\begin{minipage}[t]{0.46\linewidth}
\vskip 0pt
\centering
\captionsetup{type=table}
\caption{Variant study on different visual imagination targets and PVI head design.}
\label{tab:target_feature_ablation}
\small
\setlength{\tabcolsep}{3.5pt}
\resizebox{\linewidth}{!}{
\begin{tabular}{lcccc}
\toprule
Variant & NE $\downarrow$ & SR $\uparrow$ & OSR $\uparrow$ & SPL $\uparrow$ \\
\midrule
Front view only    & 62.94 & 50.07 & 61.99 & 44.15 \\
Center crop        & 69.06 & 47.25 & 59.59 & 41.86 \\
Future full frame  & 55.76 & 47.25 & 64.60 & 41.66 \\
Future center crop & 79.15 & 42.67 & 56.42 & 37.65 \\
MLP                & 56.29 & 46.76 & 64.81 & 41.31 \\
ProtoVis-Nav       & \textbf{51.77} & \textbf{50.78} & \textbf{66.50} & \textbf{44.68} \\
\bottomrule
\end{tabular}
}
\end{minipage}
\hfill
\begin{minipage}[t]{0.52\linewidth}
\vskip 0pt
\centering
\includegraphics[width=\linewidth]{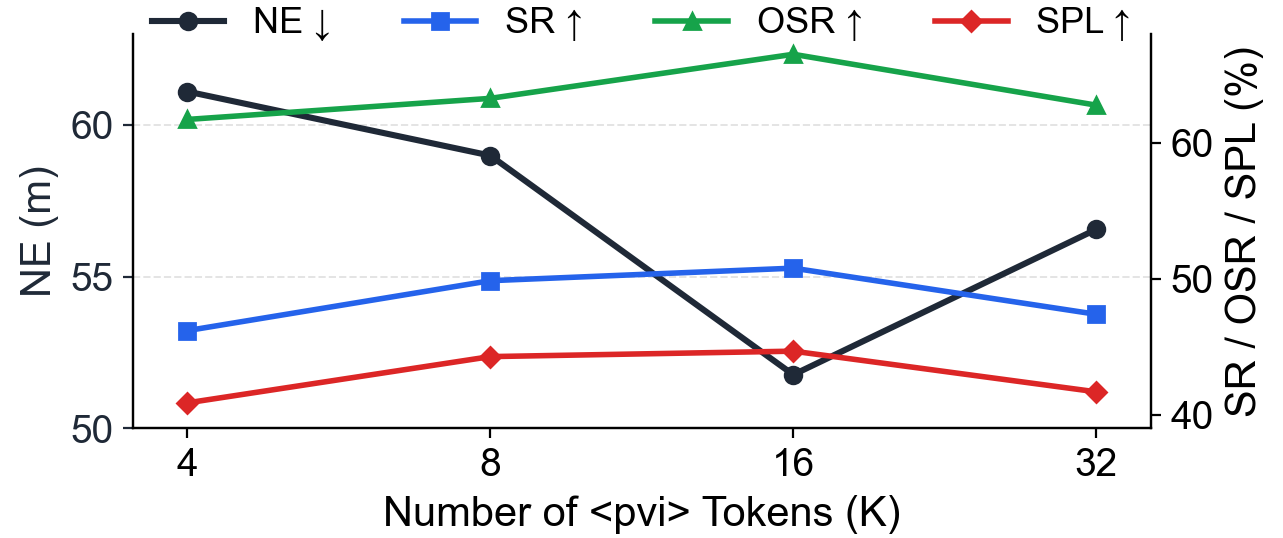}
\caption{Effect of PVI token count.}
\label{fig:pvi_token_ablation}
\end{minipage}

\end{figure}

\subsection{Variant Study}
\label{sec:ablation}

We conduct variant studies on the Test Seen split to analyze three design choices of ProtoVis-Nav: the visual imagination target, the PVI head design, and the number of \texttt{<pvi>} tokens.

\textbf{Visual imagination target.}
We first evaluate different choices of the visual imagination target. 
\textit{Front view only} always extracts the target-region feature from the current front-view image, without adaptive selection between front and down views. 
\textit{Center crop} uses the image center as the crop location, instead of projecting the next intended UAV location onto the current image plane. 
\textit{Future full frame} uses the visual feature of the full future observation, while \textit{Future center crop} uses the center crop of the future observation. 
As shown in Table~\ref{tab:target_feature_ablation}, ProtoVis-Nav achieves the best overall performance. 
Compared with \textit{Center crop}, ProtoVis-Nav reduces NE from $69.06$m to $51.77$m, showing that the projected target region is more informative than a fixed image-center target. 
Compared with \textit{Future full frame} and \textit{Future center crop}, ProtoVis-Nav also performs better, suggesting that focusing on the localized region around the next intended UAV location offers a more direct supervision target for aligning visual context with waypoint prediction.

\textbf{PVI head design.}
We then study the design of the visual imagination head. 
The \textit{MLP} variant directly regresses the target-region feature with an MLP, while ProtoVis-Nav predicts feature prototypes and linearly combines them to estimate the target feature. 
As shown in Table~\ref{tab:target_feature_ablation}, ProtoVis-Nav outperforms the MLP variant across all metrics. 
This suggests that the goal of visual imagination is not merely to fit the target feature with a high-capacity output head. 
Instead, it is more important to force the \texttt{<pvi>} tokens to encode visual cues that are useful for waypoint prediction. 
By forming the prediction as a linear combination of \texttt{<pvi>} tokens, the prototype-based design makes the imagination objective act more directly on the token representations used by the \texttt{<wp>} token.

\textbf{Number of \texttt{<pvi>} tokens.}
We analyze the effect of the number of PVI tokens. 
The value of $K$ controls the number of visual prototypes for representing the imagined target-region feature. 
We compare $K=\{4,8,16,32\}$ to study how prototype capacity affects navigation performance. 
As shown in Fig.~\ref{fig:pvi_token_ablation}, the best overall performance is achieved at $K=16$. 
When $K$ is too small, the PVI tokens may not provide enough capacity to represent the target-region visual feature. 
When $K$ becomes too large, the target-region information may be scattered across redundant PVI tokens, making the context less compact for the \texttt{<wp>} token and weakening its connection to waypoint prediction. 
Therefore, we use $K=16$ as the default setting.

\subsection{PVI Spatial Visualization and Waypoint Token Attention}
\label{sec:appendix_pvi_visualization}

\begin{figure}[!htbp]
\centering
\begin{minipage}[t]{0.49\linewidth}
\vskip 0pt
\centering
\captionsetup{type=figure,skip=0pt,font=small,justification=raggedright,singlelinecheck=false}
\includegraphics[width=\linewidth]{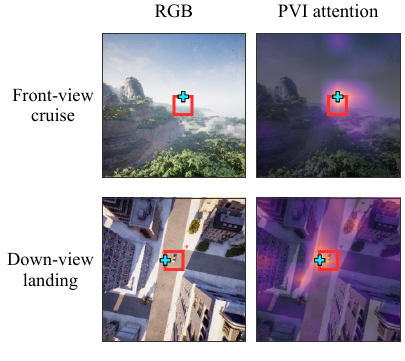}
\par\vskip 4pt
\caption{Visualization of PVI attention maps over the current images. Red boxes mark the ground-truth projected target regions, and cyan markers indicate the PVI attention peaks.}
\label{fig:pvi_spatial_examples}
\end{minipage}%
\hfill
\begin{minipage}[t]{0.49\linewidth}
\vskip 0pt
\centering
\captionsetup{type=figure,skip=0pt,font=small,justification=raggedright,singlelinecheck=false}
\includegraphics[width=\linewidth]{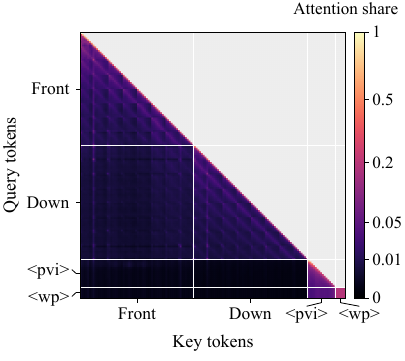}
\par\vskip 4pt
\caption{Mean causal attention over 1,000 frames. Text and UAV-state tokens are omitted; each row is renormalized over the remaining groups.}
\label{fig:pvi_token_attention}
\end{minipage}%
\end{figure}

\textbf{Spatial visualization.}
We visualize two examples by overlaying the PVI attention map on the current image (Fig.~\ref{fig:pvi_spatial_examples}). The cruise and landing examples connect the intended next UAV movement to local visual regions in the front and down views, respectively. In both cases, the attention peak lies within or near the projected target region. This suggests that PVI focuses on visual regions relevant to the next intended UAV movement.

\textbf{Waypoint token attention.}
We analyze the waypoint token's causal attention over 1,000 frames (Fig.~\ref{fig:pvi_token_attention}). We average the attention matrices across frames to summarize interactions between image, PVI, and waypoint tokens. Its attention concentrates on the PVI tokens. The waypoint token can combine this movement-related visual context with the instruction and UAV state through causal attention. Together, the visualizations and analysis suggest that PVI identifies locally action-relevant visual evidence and influences the waypoint token's representation through attention.

\subsection{Qualitative Analysis}
\label{sec:qualitative_analysis}

\begin{figure}[!htbp]
    \centering
    \includegraphics[width=0.90\linewidth]{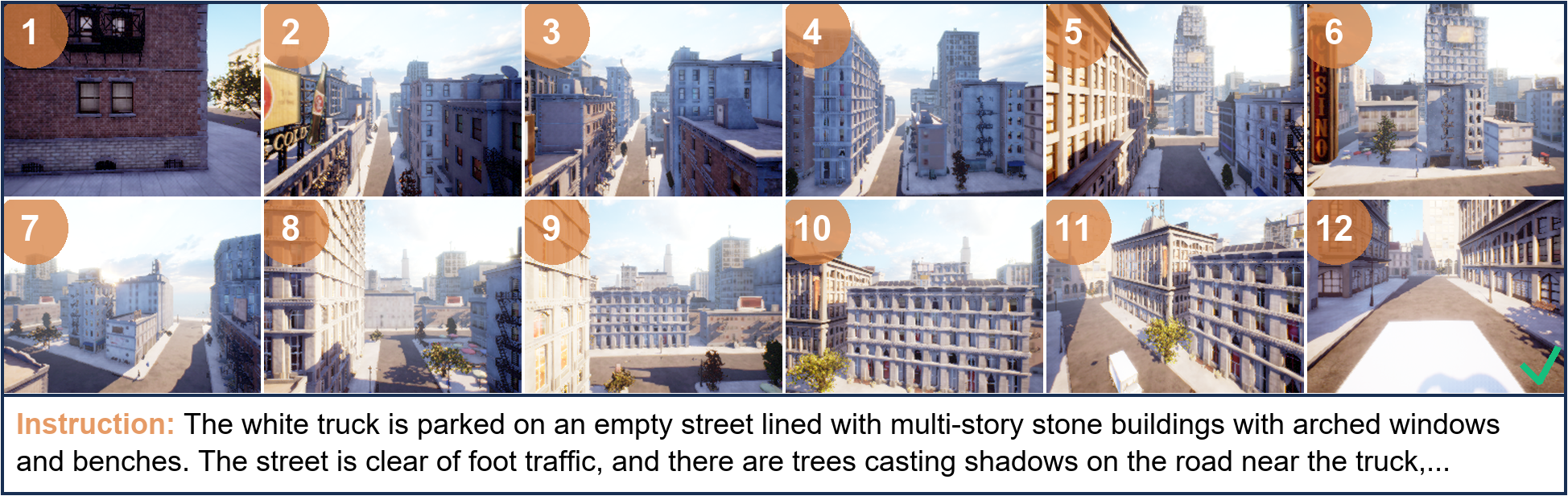}
    \par\vskip 4pt
    \captionsetup{skip=0pt}
    \caption{Qualitative example of ProtoVis-Nav on OpenUAV. The UAV follows the instruction and gradually approaches the target region. The green check mark indicates successful navigation.}
    \label{fig:qualitative}
\end{figure}

We provide qualitative examples to illustrate how ProtoVis-Nav navigates toward the target. 
As shown in Fig.~\ref{fig:qualitative}, the instruction describes a white truck parked on an empty street surrounded by multi-story stone buildings. 
In this example, the UAV gradually moves along the street, adjusts its viewpoint across consecutive observations, and finally stops near the target truck. 
The sequential observations show that ProtoVis-Nav maintains target-relevant visual context for waypoint prediction under changing viewpoints. 
This result is consistent with our motivation that prototypical visual features provide useful context for waypoint prediction.

\subsection{Limitations and Social Impact}
\label{sec:limitations}

Despite the improvements, ProtoVis-Nav still faces challenges in complex aerial environments. 
Based on our observation, most failure cases are caused by collisions, especially with trees and thin obstacles such as branches, poles, and wires. 
These obstacles often occupy only a small image region and may be partially occluded or visually ambiguous. This makes obstacle boundaries and available clearance difficult to estimate from RGB observations.
In future work, we will explore stronger obstacle-aware perception and planning mechanisms, including depth or LiDAR inputs, explicit collision-risk modeling, and anti-collision trajectory refinement for safer UAV navigation.
Our closed-loop evaluation is conducted in simulation. Future work will extend validation to real-world UAV flights and hardware-in-the-loop experiments to assess closed-loop navigation performance under physical sensing and control conditions.

ProtoVis-Nav may benefit applications such as search-and-rescue, disaster response, and infrastructure inspection by improving UAV navigation from language instructions. 
At the same time, UAV navigation systems may raise safety, privacy, and security concerns if deployed without sufficient safeguards, and should therefore be used with human oversight and deployment constraints.

\section{Conclusion}

We propose ProtoVis-Nav, a Prototypical Visual Imagination framework for UAV vision-language navigation. 
The key idea is to ground waypoint prediction in local visual context associated with the next intended movement, which is especially important for UAVs where small 3D motion errors may lead to unsafe navigation behavior. 
ProtoVis-Nav imagines an outcome-associated local visual representation for the intended next UAV movement. 
Specifically, it predicts feature prototypes and linearly combines them to estimate the target-region visual feature, and the prototype token representations provide context for waypoint prediction. 
This design promotes better alignment between visual context and waypoint outputs.
Extensive experiments on OpenUAV and HUGE-Bench demonstrate the effectiveness of ProtoVis-Nav. 

\clearpage
\endgroup

\begin{ack}
This paper was sponsored by the National Natural Science Foundation of China (No.~U25A20434).
\end{ack}

\bibliographystyle{unsrt}
\bibliography{references}

\newpage
\appendix
\section{Additional Implementation Details}

\subsection{Stop Supervision}
\label{secsec:stop head}

We design a simple stop supervision strategy to exploit the VLM's internal grounding ability for deciding when the UAV should terminate navigation. We supervise the stop head with a distance-aware soft label, encouraging the VLM to ground the stopping decision in the current visual observation and its distance to the target.

Given the output hidden state of the \texttt{<stop>} token, we predict the stopping probability with a linear layer followed by a sigmoid activation:
\begin{equation}
    p_{\mathrm{stop}}
    =
    \sigma
    \left(
    \mathrm{Linear}(\bm{h}_{stop})
    \right),
\end{equation}
where $\bm{h}_{stop}$ is the hidden state of \texttt{<stop>} and $p_{\mathrm{stop}}\in[0,1]$ is the predicted stop probability.

During training, we construct the supervision label according to the distance $d$ between the current UAV position and the target position:
\begin{equation}
    y_{\mathrm{stop}}
    =
    \sigma
    \left(
    \frac{20-d}{5}
    \right).
\end{equation}
This soft label assigns $y_{\mathrm{stop}}=0.5$ when the UAV is 20 meters away from the target. As the UAV gets closer to the target, $y_{\mathrm{stop}}$ increases toward 1; when it is farther away, the label decreases toward 0. We train the stop head with binary cross-entropy:
\begin{equation}
    \mathcal{L}_{stop}
    =
    \mathrm{BCE}
    \left(
    p_{\mathrm{stop}},
    y_{\mathrm{stop}}
    \right).
\end{equation}

During inference, the UAV sends a stop signal when:
\begin{equation}
    p_{\mathrm{stop}} \geq 0.75.
\end{equation}
This design provides a soft and distance-aware stopping objective, encouraging the model to stop only when its visual-language grounding indicates that the UAV is sufficiently close to the target.

\subsection{Waypoint Head}
\label{secsec:waypoint head}

Following TravelUAV~\cite{wang2024towards}, we use the same waypoint head design to predict a coarse waypoint from the hidden state of the \texttt{<wp>} token and the UAV kinematic context. Let $\bm{h}_{wp}\in\mathbb{R}^{B\times D}$ denote the output hidden state at the \texttt{<wp>} token position, where $B$ is the batch size and $D$ is the token hidden dimension of the VLM. Since Qwen3VL-2B-Instruct uses a hidden size of 2048, we set $D=2048$.

Besides the \texttt{<wp>} token feature, the waypoint head also uses the historical trajectory and the current UAV orientation. For each sample, the historical trajectory is represented as
\begin{equation}
    \bm{H}_i = \{\bm{p}_{i,1}, \bm{p}_{i,2}, \ldots, \bm{p}_{i,T_i}\},
    \quad
    \bm{p}_{i,t}\in\mathbb{R}^{3},
\end{equation}
where $T_i$ is the trajectory length of the $i$-th sample. Each historical position is first encoded by an MLP, and then averaged over time:
\begin{equation}
    \bm{h}_{hist}
    =
    \frac{1}{T_i}
    \sum_{t=1}^{T_i}
    \mathrm{MLP}_{hist}(\bm{p}_{i,t}),
    \quad
    \bm{h}_{hist}\in\mathbb{R}^{B\times D}.
\end{equation}
The UAV orientation $\bm{O}\in\mathbb{R}^{B\times3}$ is encoded by another MLP:
\begin{equation}
    \bm{h}_{ori}
    =
    \mathrm{MLP}_{ori}(\bm{O}),
    \quad
    \bm{h}_{ori}\in\mathbb{R}^{B\times D}.
\end{equation}
The kinematic context is obtained by summing the trajectory and orientation features:
\begin{equation}
    \bm{h}_{ctx}
    =
    \bm{h}_{hist}
    +
    \bm{h}_{ori}.
\end{equation}

We combine the \texttt{<wp>} token feature, the kinematic context, and a learnable waypoint embedding:
\begin{equation}
    \bm{h}
    =
    \bm{h}_{wp}
    +
    \bm{h}_{ctx}
    +
    \bm{e}_{wp},
\end{equation}
where $\bm{e}_{wp}\in\mathbb{R}^{D}$ is the learnable waypoint embedding. An MLP predicts the coarse waypoint:
\begin{equation}
    \hat{\bm{w}}
    =
    \mathrm{MLP}_{wp}(\bm{h}),
    \quad
    \hat{\bm{w}}\in\mathbb{R}^{B\times4}.
\end{equation}
The trajectory completion module converts the waypoint to an executable short-horizon trajectory.

\subsection{Trajectory Completion}
\label{secsec:Trajectory Completion}

\begin{figure}[t]
    \centering
    \includegraphics[width=0.6\linewidth]{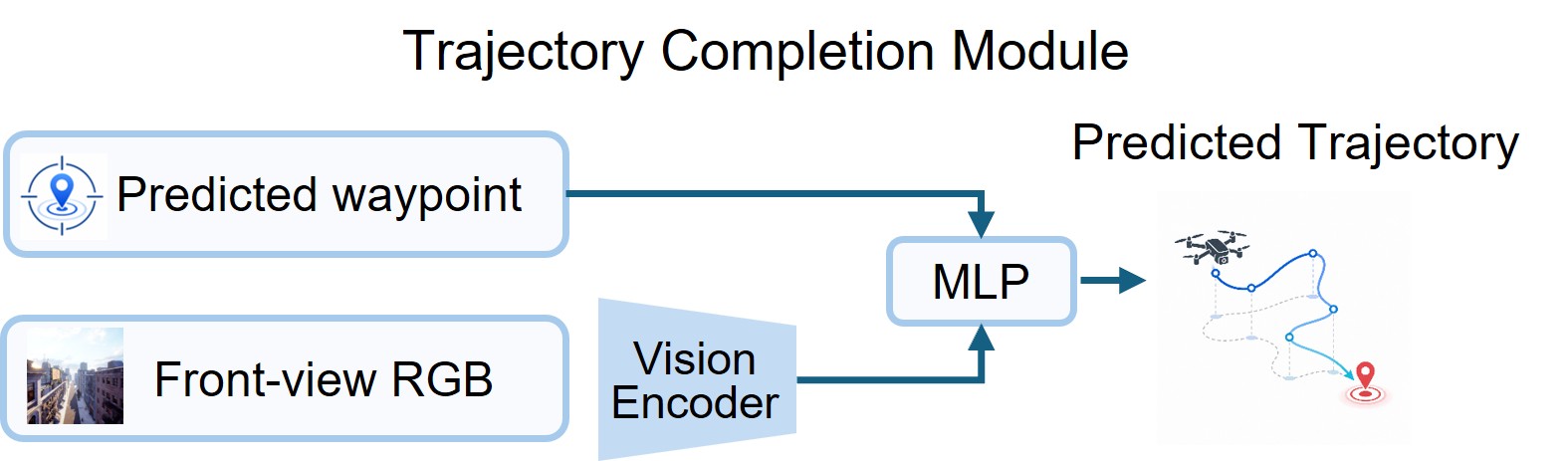}
    \caption{Illustration of the trajectory completion module. The predicted coarse waypoint and the current front-view RGB image are fused to generate a fine-grained local trajectory for UAV execution.}
    \label{fig:traj_completion}
\end{figure}

Following TravelUAV~\cite{wang2024towards}, we use the same trajectory completion module to convert a coarse waypoint into a short local trajectory, as illustrated in Fig.~\ref{fig:traj_completion}. In the dataset, each coarse waypoint is associated with a short trajectory segment. Specifically, one waypoint corresponds to 5 local trajectory steps used for execution. Therefore, after the waypoint head predicts a coarse waypoint, we use the trajectory completion module to convert this single waypoint into a short local trajectory. During training, the module is supervised to predict 7 local trajectory points for a richer short-horizon signal, while during execution we only use the first 5 predicted points to control the UAV.

The trajectory completion module can be viewed as a visual-assisted trajectory refiner: given the coarse waypoint predicted by the LLM-based waypoint head, it uses the current front-view image to complete a short local trajectory. This module does not participate in the VLM forward pass and does not read the language prompt.

Let $\bm{I}_t^{front}$ denote the current front-view image, and let $\bm{w}_{llm}\in\mathbb{R}^{3}$ denote the coarse waypoint predicted by the waypoint head and transformed into the current UAV coordinate system. We first extract visual features from the front-view image using a ViT encoder:
\begin{equation}
    \bm{V}_t = \mathrm{ViT}(\bm{I}_t^{front}).
\end{equation}
After removing the CLS token, the remaining patch features are passed through a visual MLP and then averaged to obtain a compact image feature:
\begin{equation}
    \bm{f}_{img}
    =
    \mathrm{Pool}
    \left(
    \mathrm{MLP}_{vis}(\bm{V}_t^{patch})
    \right).
\end{equation}
The image feature and the coarse waypoint are then concatenated:
\begin{equation}
    \bm{z}
    =
    [\bm{f}_{img}; \bm{w}_{llm}].
\end{equation}
Finally, an MLP predicts a sequence of local trajectory points:
\begin{equation}
    \hat{\bm{T}}
    =
    \mathrm{MLP}_{traj}(\bm{z}),
    \quad
    \hat{\bm{T}}\in\mathbb{R}^{7 \times 3}.
\end{equation}

The predicted trajectory contains 7 local points during training. At inference time, we take the first 5 points from $\hat{\bm{T}}$, transform them back to the world coordinate system, and use them for UAV execution in AirSim. This module does not re-interpret the instruction or re-localize the target. Instead, it refines the coarse waypoint into a smoother and more executable short-horizon trajectory using the current front-view visual context.

\clearpage
\section{PVI Spatial Consistency and Waypoint Accuracy}
\label{sec:appendix_pvi_alignment}

We evaluate spatial consistency on 1,000 Test Seen and 1,000 Test Unseen frames from OpenUAV, each at $256\times256$ resolution. For each frame, we compute a cosine-similarity map between the predicted PVI feature and local image-patch features, then apply soft-argmax to obtain the PVI-indicated location. PVI--GT measures its pixel distance to the projected expert target, PVI--WP measures its distance to the projected predicted waypoint, and WP--GT measures the distance between the projected predicted and expert waypoints. PVI--GT uses all 2,000 frames; comparisons involving projected waypoints use 1,959 valid frames.

\begin{table}[htbp]
\centering
\caption{Spatial alignment on OpenUAV. Values are percentages with Wilson 95\% confidence intervals in brackets.}
\label{tab:pvi_spatial_alignment}
\small
\setlength{\tabcolsep}{3pt}
\begin{tabular}{llccc}
\toprule
Model & Relation & $\leq16$ px & $\leq24$ px & $\leq32$ px \\
\midrule
\multicolumn{5}{l}{\textbf{Test Seen}} \\
ProtoVis-Nav & PVI--GT & 55.10 [52.00, 58.16] & 81.90 [79.39, 84.16] & 87.20 [84.99, 89.13] \\
 & PVI--WP & 53.93 [50.80, 57.03] & 80.29 [77.68, 82.66] & 86.11 [83.80, 88.13] \\
 & WP--GT & 82.12 [79.60, 84.40] & 87.54 [85.32, 89.46] & 92.54 [90.73, 94.03] \\
Base & WP--GT & 81.70 [79.15, 84.00] & 87.12 [84.87, 89.07] & 92.23 [90.38, 93.75] \\
\midrule
\multicolumn{5}{l}{\textbf{Test Unseen}} \\
ProtoVis-Nav & PVI--GT & 58.00 [54.92, 61.02] & 79.80 [77.20, 82.17] & 88.40 [86.27, 90.24] \\
 & PVI--WP & 57.55 [54.43, 60.61] & 81.53 [78.98, 83.84] & 90.10 [88.07, 91.82] \\
 & WP--GT & 80.51 [77.91, 82.87] & 86.43 [84.14, 88.43] & 90.92 [88.96, 92.56] \\
Base & WP--GT & 79.92 [77.30, 82.31] & 86.34 [84.05, 88.35] & 90.21 [88.19, 91.92] \\
\bottomrule
\end{tabular}
\end{table}

At the 32-pixel threshold, PVI--GT alignment reaches $87.20\%$ on Test Seen and $88.40\%$ on Test Unseen, and PVI--WP alignment reaches $86.11\%$ and $90.10\%$, respectively (Table~\ref{tab:pvi_spatial_alignment}). ProtoVis-Nav also improves WP--GT alignment over Base at all three thresholds on both splits. These results show spatial agreement among PVI, the projected target region, and the predicted waypoint.

We further measure mean waypoint prediction error on 1,000 sampled OpenUAV test frames. As shown in Table~\ref{tab:pvi_waypoint_error}, PVI reduces the mean error from $0.4630$m to $0.4576$m, corresponding to a $5.4$\,mm improvement in the continuous waypoint output.

\begin{table}[htbp]
\centering
\caption{Mean waypoint prediction error on 1,000 sampled OpenUAV test frames.}
\label{tab:pvi_waypoint_error}
\small
\setlength{\tabcolsep}{6pt}
\begin{tabular}{lc}
\toprule
Model & Mean waypoint error (m) $\downarrow$ \\
\midrule
Base & 0.4630 \\
ProtoVis-Nav & \textbf{0.4576} \\
\bottomrule
\end{tabular}
\end{table}

\clearpage
\section{PVI Feature Similarity and Retrieval}
\label{sec:appendix_pvi_feature_retrieval}

We evaluate the same 1,000 Seen and 1,000 Unseen frames used in Table~\ref{tab:pvi_spatial_alignment}. Each $256\times256$ frame is densely sampled into 841 candidate patches of size $32\times32$ with a stride of 8 pixels.

First, we measure the direct cosine similarity between the predicted PVI representation and the ground-truth target-patch feature, using a randomly selected patch as the baseline. Second, we rank all candidate patches by cosine similarity and report the first-relevant rank. A candidate is considered relevant if its center is less than 32 pixels from the ground-truth target center, within one visual-token width. We report spatial Recall@1, Recall@5, and Recall@10 under the same criterion.

\begin{table}[htbp]
\centering
\caption{Direct PVI feature similarity and spatial retrieval accuracy. Values in brackets are 95\% confidence intervals.}
\label{tab:pvi_feature_retrieval}
\small
\setlength{\tabcolsep}{4pt}
\begin{tabular}{lcc}
\toprule
Metric & Test Seen & Test Unseen \\
\midrule
PVI--GT cosine similarity & 0.8440 [0.8402, 0.8477] & 0.8449 [0.8415, 0.8482] \\
PVI--random cosine similarity & 0.7666 [0.7604, 0.7728] & 0.7535 [0.7464, 0.7605] \\
Mean first-relevant rank ($<32$ px) $\downarrow$ & 11.19 [9.22, 13.57] & 15.56 [13.38, 18.05] \\
Recall@1 ($<32$ px) & 27.00\% [24.34\%, 29.84\%] & 21.10\% [18.68\%, 23.74\%] \\
Recall@5 ($<32$ px) & 63.50\% [60.47\%, 66.43\%] & 54.90\% [51.80\%, 57.96\%] \\
Recall@10 ($<32$ px) & 80.40\% [77.83\%, 82.74\%] & 69.30\% [66.37\%, 72.08\%] \\
\bottomrule
\end{tabular}
\end{table}

As shown in Table~\ref{tab:pvi_feature_retrieval}, the PVI--target cosine similarity reaches $0.8440$ on Seen and $0.8449$ on Unseen, compared with $0.7666$ and $0.7535$ for random patches. Among 841 candidate patches, the mean first-relevant rank is $11.19$ on Seen and $15.56$ on Unseen. At the 32-pixel threshold, Recall@1/5/10 reaches $27.00\%$/$63.50\%$/$80.40\%$ on Seen and $21.10\%$/$54.90\%$/$69.30\%$ on Unseen. These results show that the PVI representation contains information about the ground-truth target patch.

\section{Robustness to Observation and Execution Perturbations}
\label{sec:appendix_execution_perturbations}

We evaluate ProtoVis-Nav on OpenUAV Test Seen under three perturbations. Camera image noise affects the visual inputs, pose estimation noise affects the position and yaw estimates supplied to the model, and execution mismatch introduces action execution errors and horizontal drift.

\begin{table}[htbp]
\centering
\caption{Robustness of ProtoVis-Nav to observation and execution perturbations on OpenUAV Test Seen.}
\label{tab:execution_perturbations}
\small
\setlength{\tabcolsep}{6pt}
\begin{tabular}{lcccc}
\toprule
Variant & NE (m) $\downarrow$ & SR (\%) $\uparrow$ & OSR (\%) $\uparrow$ & SPL (\%) $\uparrow$ \\
\midrule
ProtoVis-Nav & \textbf{51.77} & \textbf{50.78} & \textbf{66.50} & \textbf{44.68} \\
+ Camera image noise & 52.33 & 49.51 & 64.80 & 44.17 \\
+ Position and yaw noise & 56.24 & 46.42 & 63.11 & 41.25 \\
+ Execution mismatch & 59.92 & 47.69 & 62.41 & 42.02 \\
\bottomrule
\end{tabular}
\end{table}

As shown in Table~\ref{tab:execution_perturbations}, camera image noise has the smallest effect on SR and SPL, reducing them from $50.78\%$ and $44.68\%$ to $49.51\%$ and $44.17\%$. Pose estimation noise produces the lowest SR ($46.42\%$), and execution mismatch produces the highest NE ($59.92$m), illustrating the effect of state estimation and execution accuracy on closed-loop navigation.

\clearpage
\section{Evaluation on UAV-Flow}
\label{sec:appendix_uav_flow}

We compare ProtoVis-Nav with Base on UAV-Flow \cite{wang2025uavflow}. Normalized Dynamic Time Warping (nDTW) measures trajectory alignment. We use double-blind human annotation to determine the success rate based on manual inspection of whether the trajectory semantically satisfies the instruction, consistent with official UAV-Flow settings.

\begin{table}[htbp]
\centering
\caption{Closed-loop results on UAV-Flow.}
\label{tab:uav_flow}
\small
\setlength{\tabcolsep}{6pt}
\begin{tabular}{lcccc}
\toprule
Model & Trajectories & nDTW $\uparrow$ & Successes & SR (\%) $\uparrow$ \\
\midrule
Base & 273 & 0.2971 & 185 & 67.77 \\
ProtoVis-Nav & 273 & \textbf{0.3104} & \textbf{197} & \textbf{72.16} \\
\bottomrule
\end{tabular}
\end{table}

On the full test split of UAV-Flow, ProtoVis-Nav improves closed-loop nDTW from $0.2971$ to $0.3104$ and SR from $67.77\%$ to $72.16\%$ over Base (Table~\ref{tab:uav_flow}), indicating closer alignment with the reference trajectories and higher task success. UAV-Flow evaluates fine-grained flight skills, so these gains suggest that target-region visual features also provide useful context for short-range motion control.

\section{Backbone Comparison}
\label{sec:appendix_backbone_comparison}

We compare Qwen3-VL and Wan2.1-VACE in a controlled backbone-transfer experiment on OpenUAV Test Seen, using the same PVI design, waypoint heads, action representation, training data, and evaluation protocol. Inference latency is measured for a single forward pass on the same GPU.

\begin{table}[htbp]
\centering
\caption{Navigation performance and inference latency of different backbones on OpenUAV Test Seen.}
\label{tab:backbone_comparison}
\small
\setlength{\tabcolsep}{5pt}
\begin{tabular}{lccccc}
\toprule
Backbone & NE (m) $\downarrow$ & SR (\%) $\uparrow$ & OSR (\%) $\uparrow$ & SPL (\%) $\uparrow$ & Latency (ms) $\downarrow$ \\
\midrule
Wan2.1-VACE & 111.46 & 17.25 & 22.30 & 14.79 & 333.83 \\
Qwen3-VL & \textbf{51.77} & \textbf{50.78} & \textbf{66.50} & \textbf{44.68} & \textbf{185.48} \\
\bottomrule
\end{tabular}
\end{table}

As shown in Table~\ref{tab:backbone_comparison}, Qwen3-VL increases SR from $17.25\%$ to $50.78\%$ and SPL from $14.79\%$ to $44.68\%$, while reducing latency from $333.83$ms to $185.48$ms. World action models (WAMs) offer explicit physical foresight for longer-horizon planning, with diffusion-based generation incurring iterative inference costs. ProtoVis-Nav trades this multi-step foresight for efficient local waypoint prediction, conditioning the next waypoint on an outcome-associated local visual representation in a single forward pass.

\clearpage
\section{Effect of inference-time PVI interventions}
\label{sec:appendix_pvi_interventions}
We further examine the contribution of PVI context through inference-time interventions. We evaluate four PVI intervention settings on Test Seen Full. \textit{Intact} uses the PVI representations predicted for the current sample; \textit{Shuffled} exchanges these representations across samples; \textit{Masked} blocks PVI information from reaching the waypoint predictor; and \textit{Random} replaces it with random features. As shown in Table~\ref{tab:pvi_interventions}, Intact achieves the highest SR, OSR, and SPL and the lowest NE across these settings. Shuffling reduces SR from $50.78\%$ to $41.09\%$, while masking and random replacement further reduce it to $37.87\%$ and $36.75\%$, respectively. These results show that the waypoint predictor uses sample-specific PVI information during inference, with coherent local visual context contributing to successful navigation.

\begin{table}[!htbp]
\centering
\captionsetup{skip=0pt}
\caption{Effect of inference-time PVI interventions on OpenUAV Test Seen Full.}
\label{tab:pvi_interventions}
\vskip 4pt
\small
\setlength{\tabcolsep}{5pt}
\begin{tabular}{llcccc}
\toprule
Model & PVI setting & NE (m) $\downarrow$ & SR (\%) $\uparrow$ & OSR (\%) $\uparrow$ & SPL (\%) $\uparrow$ \\
\midrule
Base & -- & 55.67 & 46.47 & 62.69 & 41.14 \\
\midrule
\multirow{4}{*}{ProtoVis-Nav}
& Intact & \textbf{51.77} & \textbf{50.78} & \textbf{66.50} & \textbf{44.68} \\
& Shuffled & 57.36 & 41.09 & 54.56 & 36.88 \\
& Masked & 70.99 & 37.87 & 49.65 & 34.74 \\
& Random & 77.18 & 36.75 & 47.69 & 33.38 \\
\bottomrule
\end{tabular}
\end{table}

\end{document}